%% file: catt.tex
\documentclass{article}
\usepackage{spconf,amsmath,graphicx}
\usepackage{amsfonts}
\usepackage{comment}
\usepackage{cite}

\usepackage{microtype}
\usepackage{graphicx}
\usepackage{amsmath}
\usepackage{amssymb}
\usepackage{graphicx}
\usepackage{xcolor}
\usepackage[normalem]{ulem}
\usepackage{multirow}
\usepackage{booktabs}

\title{Context-Aware Transformer Transducer \\for Speech Recognition}
\name{\begin{tabular}{c}Feng-Ju Chang\textsuperscript{\rm *}, Jing Liu\textsuperscript{\rm *}\thanks{*Joint First Authors}, Martin Radfar, Athanasios Mouchtaris, \\ Maurizio Omologo, Ariya Rastrow, Siegfried Kunzmann\end{tabular}}

\address{Amazon Alexa \\ 
{\small \tt \{fengjc,jlmk,radfarmr,mouchta,omologo,arastrow,kunzman\}@amazon.com}}

\begin{document}
%
\maketitle
%
\input{abstract.tex}
%
\input{introduction.tex}

\input{proposed_approach.tex}

\input{Experiments.tex}

\input{results_and_discussions.tex}

\input{conclusion.tex}

\input{acknowledgement.tex}


\bibliographystyle{IEEEbib}
\bibliography{refs,strings}

\end{document}

%% file: abstract.tex
\begin{abstract}

End-to-end (E2E) automatic speech recognition (ASR) systems often have difficulty recognizing uncommon words, that appear infrequently in the training data. One promising method, to improve the recognition accuracy on such rare words, is to latch onto personalized/contextual information at inference. In this work, we present a novel context-aware transformer transducer (CATT) network that improves the state-of-the-art transformer-based ASR system by taking advantage of such contextual signals. Specifically, we propose a multi-head attention-based context-biasing network, which is jointly trained with the rest of the ASR sub-networks. We explore different techniques to encode contextual data and to create the final attention context vectors. We also leverage both BLSTM and pretrained BERT based models to encode contextual data and guide the network training. Using an in-house far-field dataset, we show that CATT, using a BERT based context encoder, improves the word error rate of the baseline transformer transducer and outperforms an existing deep contextual model by 24.2\% and 19.4\% respectively.

\end{abstract}

%% file: introduction.tex
\section{Introduction}
\label{sec:intro}
E2E ASR systems such as connectionist temporal classification (CTC) \cite{Graves2006ConnectionistTC}, listen-attend-spell (LAS) \cite{LAS2016}, recurrent neural network transducer (RNN-T) \cite{graves2012sequence} and transformer \cite{Dong2018SpeechTransformerAN} have gained significant interest due to their superior performance over hybrid HMM-DNN systems, as sufficient training data is available \cite{chiu2018state}. While hybrid models optimize the acoustic model (AM), pronunciation model (PM) and language model (LM) independently, E2E models implicitly subsume these modules and jointly optimize them to output word sequences \cite{sennrich-etal-2016-neural, kudo2018subword} directly given an input sequence. In addition, E2E models simplify the inference pipeline without external alignment modules and LMs, which make them more capable for on-device deployment \cite{googleRNNT}.

However, one major limitation of an E2E ASR system is that it cannot accurately recognize words that appear rarely in the paired audio-to-text training data, such as entity names or personalized words \cite{pundak2018deep, chen2019joint, bruguier2019phoebe}. To address this issue, previous work has leveraged context where rare words would appear more frequently or are associated with weights, e.g. the weighted finite state transducer (WFST)~\cite{Mohri2002WeightedFT} constructed from the speaker's context \cite{williams2018contextual}, domain~\cite{liu2021domainaware}, text metadata of video~\cite{jain2020contextual,liu2020contextualizing}, dialogue state, location, or personalized information about the speaker (e.g., personalized device names or contact names)~\cite{pundak2018deep, gourav2021personalization}, and so on. 
\begin{figure*}[t]
  \centering
  \includegraphics[width=0.8\linewidth, height=6cm]{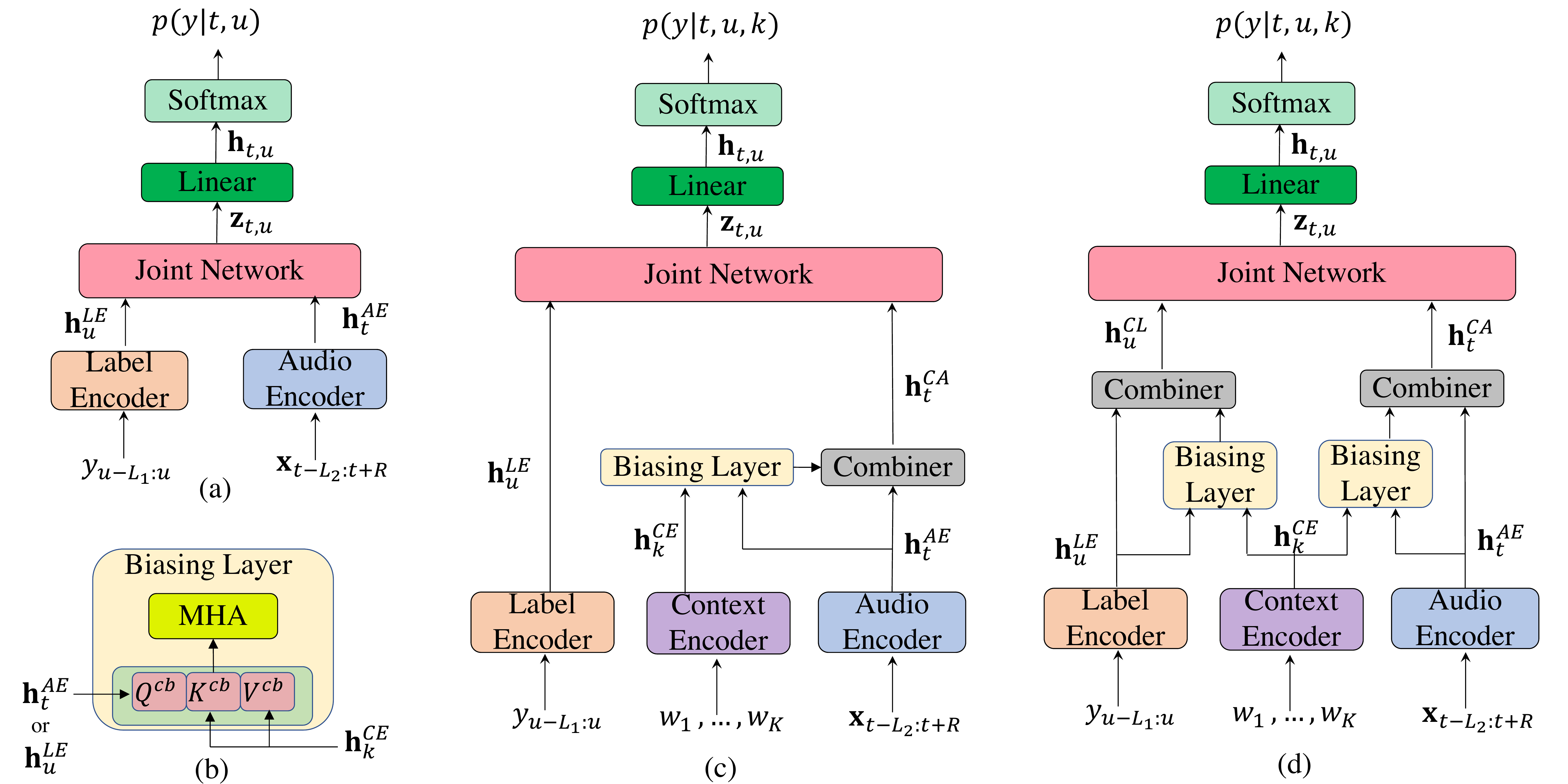}
  \caption{(a) Transformer Transducer (b) The proposed context biasing layer via multi-head attention (MHA) between audio/label and context embeddings. The proposed context-aware transformer transducer (CATT) with (c) audio embeddings or (d) audio+label embeddings to attend the context embeddings}
  \label{fig:catt_diagram}
\end{figure*}

In general, the methods that integrate the above context into E2E ASR systems can be divided into two categories: post-training integration \cite{williams2018contextual,gourav2021personalization,zhao2019shallow,aleksic2015bringing,gulcehre2015using,huang2020class,Chen2019EndtoendCS,Toshniwal2018ACO,Kannan2018AnAO} and during-training integration \cite{pundak2018deep,jain2020contextual,sriram2017cold}. The former is only applied in inference time while the latter occurs in both training and inference. Shallow fusion \cite{zhao2019shallow} is one of the dominant paradigms used to incorporate WFST via an independently-trained on-the-fly rescoring framework to adjust the LM weights of $n$-grams relevant to the particular context \cite{aleksic2015bringing}. Another post-training integration method is deep fusion \cite{gulcehre2015using}, which integrates the external LM into the E2E ASR by fusing together the hidden states of the external LM and the decoder. A major drawback of post-training integration, however, is that it requires external LMs to rescore the ASR model's outputs and it is sensitive to rescoring weights.

The most relevant work to ours in the during-training integration category is contextual LAS (C-LAS) \cite{pundak2018deep}, which proposes an additional bias encoder with a location-aware attention mechanism \cite{chorowski2015attention} on top of LAS \cite{LAS2016} in order to rescore the personalized words at training and inference, with label embeddings. Similarly, contextual RNN-T (C-RNN-T) \cite{jain2020contextual} applies the same attention mechanism but using RNN-T \cite{graves2012sequence}. 
Phonetic information is explored in \cite{bruguier2019phoebe} and \cite{Chen2019JointGA} to further improve C-LAS. 


Witnessing the transformer network and its variant, transformer transducer \cite{NIPS2017_3f5ee243,tian2019self,yeh2019transformer,zhang2020transformer}, have become state-of-the-art ASR models, we propose a novel Context-Aware Transformer Transducer (CATT) network, enabling transformer transducer models to leverage contextual information both in training and inference to improve ASR performance. 

Different from C-LAS~\cite{pundak2018deep} and C-RNN-T~\cite{jain2020contextual}, we encode the contextual data not only with BLSTM~\cite{hochreiter1997long,biRNN1997}, but also with a pre-trained BERT~\cite{devlin2019bert,turc2019} model, which brings strong semantic knowledge to guide the network at training and inference. 
In addition, we propose a multi-head attention based context biasing module in order to measure the importance of the contextual phrases. 
We employ audio embeddings alone or with label embeddings to measure the importance of context, and hence create the relevant context vectors that are fed into the ASR frame by frame to improve the alignment learning.



Using an in-house far-field dataset, we demonstrate that injecting context into audio embeddings in our CATT model outperforms the baseline methods with no context, shallow fusion~\cite{zhao2019shallow}, and C-LAS~\cite{pundak2018deep}. Moreover, employing a BERT based context encoder achieves the greatest improvements. Finally, we find that injecting context into both audio and label embeddings can further improve the CATT model over the one with context injected in audio embeddings only.

%% file: proposed_approach.tex
\section{Proposed Approach}
In this section, we describe the transformer transducer model and our design of the context-aware module.

\subsection{Transformer Transducer}
\label{sec:tt}
The transformer transducer model \cite{tian2019self,yeh2019transformer,zhang2020transformer} outputs a probability distribution over output sequences $\mathbf{y}$ given input audio frames $\mathbf{x}$. The model contains three main modules (Fig.~\ref{fig:catt_diagram} (a)): an audio encoder, a label encoder, and a joint network. 

The audio encoder, $f^{\text{enc}}$, is comprised of stacked self-attention transformer layers \cite{NIPS2017_3f5ee243} that uses audio features $\mathbf{x}$ in a predefined window centered at frame $t$, [$t$-$L_2$:$t$+$R$], to produce the audio embedding $\mathbf{h}_t^{AE}$ at frame $t$,
\begin{equation}
\mathbf{h}_t^{AE}= f^{\text{enc}} (\mathbf{x}_{t-L_2:t+R})
\label{h_ae}
\end{equation}
where $\mathbf{h}_t^{AE} \in \mathbb{R}^{d_a \times 1}$ and $d_a$ is the embedding size. It is similar to the AM in a hybrid ASR system. 

The label encoder, $f^{\text{pred}}$, is also a stacked transformer network, which uses the previous $L_1$ non-blank tokens $y$ to generate the label embedding $\mathbf{h}_u^{LE}$. We use subwords as tokens.

\begin{equation}
\mathbf{h}_u^{LE}= f^{\text{pred}} (y_{u-L_1:u})
\label{h_le}
\end{equation}
where $\mathbf{h}_u^{LE} \in \mathbb{R}^{d_l \times 1}$ and $d_l$ is the embedding size; $f^{\text{pred}}$ acts similarly to the LM in a hybrid ASR system.

The joint network combines the audio encoder outputs and label encoder outputs to produce a new embedding:

\begin{equation}
\mathbf{z}_{t,u} = \phi(U \mathbf{h}_t^{AE} + V \mathbf{h}_u^{LE} + \mathbf{b}_1)
\label{joint}
\end{equation}
where $U$, $V$, $\mathbf{b}_1$ are learnable parameters to project the audio and label embeddings into the same dimension. $\phi$ is a nonlinear function; we opt for $\tanh$ in this work. $\mathbf{z}_{t,u}$ is then fed into a linear layer and Softmax layer to produce a probability distribution over output labels plus a blank symbol. 
\begin{align}
& \mathbf{h}_{t,u}=W\mathbf{z}_{t,u} + \mathbf{b}_2 \nonumber \\
& p(y|t,u) = \text{Softmax}(\mathbf{h}_{t,u}) \label{softmax}
\end{align}
where $W$ and $\mathbf{b}_2$ are trainable parameters.
When the joint network predicts a blank symbol, the model proceeds to the audio encoder ouput of the next time frame; when a non-blank symbol is predicted, the label encoder output will be updated. This results in various alignment paths, and the sum of probabilities of them provides the probability of an output sequence (with non-blank outputs) given an input sequence.

\subsection{Context-Aware Transformer Transducer (CATT)}
\label{sec:catt}
To inject context information, we modify the base transformer transducer described in Section~\ref{sec:tt}  and add two additional components: (1) a context encoder, and (2) a multi-head attention-based context biasing layer, as shown in Fig.~\ref{fig:catt_diagram} (b). 

\textbf{Context Encoder}:
The context we employ in this work contains personalized information provided by the speakers, such as speaker-defined device names, device settings, and device locations as presented in Table~\ref{train_data_context}. Each contextual word or phrase $w_k$ is first represented as subwords and then fed into the context encoder $f^{\text{context}}$, to produce fixed dimensional vector representations.
\begin{equation}\label{eq:context_embedding}
\mathbf{h}_k^{CE}= f^{\text{context}} (w_k).
\end{equation}
where $\mathbf{h}_k^{CE} \in \mathbb{R}^{d_c \times 1}$. \\
In particular, we investigate two types of context encoders, $f^{\text{context}}$:
\begin{itemize}
\item \emph{BLSTM based context embedding}: The inputs for this embedding extractor are tokenized contextual phrases by a subword tokenizer \cite{sennrich-etal-2016-neural}. Then, the last state of BLSTM is used as the embedding of a context word or phrase. Note that we train the BLSTM from scratch along with the rest of the networks.
\item \emph{BERT \cite{devlin2019bert} based context embedding}: This context encoder is pre-trained with a substantial amount of text data. Using BERT, we bring strong semantic prior knowledge to guide the training of the rest of the networks in our model. Specifically, the SmallBERT \cite{turc2019} model is employed. We investigate the case when this encoder is frozen, and only the rest of our network weights are updated. 

\end{itemize}

Note that the transformer transducer described in Section~\ref{sec:tt} exploits only the outputs of the audio encoder, $\mathbf{h}_t^{AE}$, and label encoder, $\mathbf{h}_u^{LE}$, to produce the probabilities over tokens, $p(y|t,u) = p(y|\mathbf{h}_t^{AE},\mathbf{h}_u^{LE})$. In contrast, for our context-aware transformer transducer, the output probability is conditionally dependent on the contextual data as well. Namely,  $p(y|t,u)$  becomes 
\begin{equation}
p(y|t,u,k) = p(y|\mathbf{h}_t^{AE},\mathbf{h}_u^{LE},\mathbf{h}_k^{CE}).
\end{equation}


\textbf{Multi-Head Attention based Context Biasing Layer}:
Given the context embeddings, $\mathbf{h}_k^{CE}$, from Eqn.(\ref{eq:context_embedding}), this module is designed to learn context phrase relevance to an utterance. In this way, the model can pay more attention to the frames corresponding to the entity names or personalized words to help improve their prediction accuracy. Since our base model is transformer-based \cite{NIPS2017_3f5ee243}, multi-head attention (MHA) becomes a natural choice to learn the relationships between the context embeddings and the utterance's embeddings. We first investigate the audio embeddings as queries to attend context (See Fig.~\ref{fig:catt_diagram} (c)) because the audio encoder used in this work is bi-directional transformer (i.e. the attention is computed on both previous and future frames), which covers more information than the label encoder (uni-directional transformer) about the input utterance. We also investigate using both audio embeddings and label embeddings as queries to attend context, as shown in Fig.~\ref{fig:catt_diagram} (d).

For clarity, we will omit the MHA here and focus on how we integrate context and audio embeddings with the attention mechanism. The same formulation can be applied to label embeddings as well. Specifically, we create the query, key, and value as follows (Fig.~\ref{fig:catt_diagram} (b)):    
\begin{align}
Q^{cb} = \sigma\left(X W^{cb,q} + \textbf{1} (\textbf{b}^{cb,q}_i)^\top \right) \nonumber \\
K^{cb} = \sigma\left(C W^{cb,k} + \textbf{1} (\textbf{b}^{cb,k}_i)^\top \right)  \label{cbe} \\
V^{cb} = \sigma\left(C W^{cb,v} + \textbf{1} (\textbf{b}^{cb,v}_i)^\top \right) \nonumber
\end{align}
Here, $X = [\mathbf{h}_1^{AE},\dots,\mathbf{h}_\mathcal{T}^{AE}]^{\top} \in \mathbb{R}^{\mathcal{T} \times d_a}$ is the audio encoder outputs of an utterance and $\mathcal{T}$ is the number of audio frames. $C = [\mathbf{h}_1^{CE},\dots,\mathbf{h}_K^{CE}]^{\top} \in \mathbb{R}^{K \times d_c}$ is the associated context embeddings, where $K$ is the number of context phrases. $\sigma(\cdot)$ is an activation function. $W^{cb,q} \in \mathbb{R}^{d_a \times d}$, $W^{cb,k} \in \mathbb{R}^{d_c \times d}$, $W^{cb,v} \in \mathbb{R}^{d_c \times d}$, and $\textbf{b}^{cb,*} \in \mathbb{R}^{d \times 1}$ are all learnable weight and bias parameters, and $\textbf{1} \in \mathbb{R}^{\mathcal{T} \times 1}$ is an all-ones vector.

The cross-attention between audio embeddings and context embeddings is then computed by
\begin{align}
    H^{cb} = \text{Softmax}\left(\frac{Q^{cb} (K^{cb})^\top}{\sqrt{d}}\right) V^{cb},
    \label{eq:cc}
\end{align}
where the scaling $\frac{1}{\sqrt{d}}$ is for numerical stability~\cite{NIPS2017_3f5ee243}. The resulting attention scores (Softmax part of Eqn.(\ref{eq:cc})) are used to take the weighted sum of context embeddings resulting in the context-aware matrix $H^{cb} \in \mathbb{R}^{\mathcal{T} \times d}$.  This context-aware matrix is then fused with the audio embedding matrix through a combiner, which consists of LayerNorm layers, a concatenation and a feed-forward projection layer in order to match the dimension of the label encoder outputs (Fig.~\ref{fig:catt_diagram} (c)). This context integration process can be described as
\begin{align}
& H^{concat} = [\text{LayerNorm}(X), \text{LayerNorm}(H^{cb})]  \nonumber \\
& H^{CA} = \text{FeedForward}(H^{concat}),
\label{combiner}
\end{align}
where $H^{concat} \in \mathbb{R}^{\mathcal{T} \times (d_a+d)}$ and $H^{CA} = [\mathbf{h}^{CA}_1,\dots,\mathbf{h}^{CA}_{\mathcal{T}}] \in \mathbb{R}^{\mathcal{T} \times d_{ca}}$. We finally feed the context-aware audio embeddings along with the label embeddings into the joint network. 
Similarly, when we use the label embeddings to attend context, the inputs used to compute $Q^{cb}$ become the label encoder outputs $Y = [\mathbf{h}_1^{LE},\dots,\mathbf{h}_\mathcal{U}^{LE}]^{\top} \in \mathbb{R}^{\mathcal{U} \times d_l}$ of an utterance and $\mathcal{U}$ is the number of tokens that have been predicted.

Fig.~\ref{fig:catt_diagram} (d) illustrates how we use both audio embeddings and label embeddings to attend the context. Specifically, we have two biasing layers, one taking the audio embeddings as queries, and the other taking the label embeddings as queries to evaluate the importance of context. Each of the context-aware matrices from biasing layers will be fused with the original audio or label embeddings via a combiner (Eqn.(\ref{combiner})). The context-injected audio ($\mathbf{h}_t^{CA}$) and label ($\mathbf{h}_u^{CL}$) embeddings are then fed into the joint network for alignment learning.

%% file: Experiments.tex
\section{Experimental Setup}

\subsection{Dataset}
\label{sec:dataset2}
To evaluate the proposed CATT model, we use 1,300 hours of speech utterances from our in-house de-identified far-field dataset. The amount of training set, dev set, and test set are 910 hours, 195 hours, and 195 hours respectively. We train our model on the train split and report the results on both dev and test splits. The device-directed speech data is captured using a smart speaker, and the contextual words/phrases are provided by the speakers consisting of personalized device names, entity names, device settings, and device locations, as shown in Table~\ref{train_data_context}. 


We denote the contextual phrases which are present in the transcription as relevant context, and those that are absent in the transcription as less relevant context. During training in each batch, we use all relevant contextual phrases and cap the maximum number of contextual entities at $K$=100. If the number of relevant contextual phrases is less than 100, we randomly add less relevant context until we reach a total of 100. 
In this way, every training utterance would be paired with a variety of context in each training step. This samplings allows flexibility at inference, as the model does not make any assumption about what contextual phrases will be used. Note that we assume the relevant contextual phrases are always present during inference.

During evaluation, we further divide the dev and test sets into two categories, \emph{Personalized} and \emph{Common} sets. An utterance is labeled as \emph{Personalized} if there is any word in the transcription that also appears in the contextual phrase list of personalized device names or entity names.  Otherwise, it is categorized as \emph{Common}\footnote{The utterances in the \emph{Common} set may contain contextual phrases like the device location or/and device setting.}. 
Example utterances are illustrated in Table~\ref{data_sample} and we present results on these two sets in Section~\ref{sec:results}.

\begin{table}[t]
\caption{The types and examples of contextual words/phrases}
\vskip 3pt
\centering
\label{train_data_context}
\begin{tabular}{|l|l|l|l|}
\hline
\textbf{Context Type}        & \textbf{Example Contextual Phrases}                        \\ \hline
Personalized  & baine's room, \\
Device Name  & grogu's second tv  \\\hline
Named entity & foxtel, xbox \\\hline
Device Setting & turn on, off, dim  \\ \hline
Device Location & living room, basement, hallway \\ \hline
\end{tabular}
\end{table}

\begin{table}[t]
\caption{Example data in \emph{Personalized} and \emph{Common} sets}
\vskip 3pt
\centering
\label{data_sample}
\begin{tabular}{|l|l|l|l|}
\hline
\textbf{Data Split}        & \textbf{Example}                        \\ \hline
\emph{Personalized}  & turn off baine's second TV  \\\hline
$\emph{Common}^{1}$ & dim living room light thirty percent \\\hline
\end{tabular}
\end{table}

\begin{table}[t]
\footnotesize
\caption{WERRs for 14M CATT vs.T-T and SF}
\vskip 3pt
\centering
\label{tab:14M_auto_WER}
\begin{tabular}{|l|c|c|c|c|}
\hline
  & \multicolumn{2}{c|}{\emph{Personalized}}                        & \multicolumn{2}{c|}{\emph{Common}}                          \\ \hline
WERR (\%)    & \multicolumn{1}{l|}{dev} & \multicolumn{1}{l|}{test} & \multicolumn{1}{l|}{dev} & \multicolumn{1}{l|}{test} \\ \hline
T-T                     & 0 & 0 & 0 & 0  \\ \hline
T-T + SF                & 2.9     & 1.6     & 1.0     & 2.2     \\ \hline \hline
CATT (BLSTM-CE)    & 1.4 & 1.6 & 2.1  & 2.2 \\ \hline
CATT (BLSTM-CE) + SF    & 2.9     &  3.1    & 4.1     &  4.3    \\ \hline \hline
CATT (SmallBERT-CE)    &  4.3   &  4.7     & 4.1     & 4.3     \\ \hline
CATT (SmallBERT-CE) + SF & 5.7     & 6.3     &  5.2    &    6.5   \\ \hline
\end{tabular}
\end{table}

\subsection{Model Configurations}
The proposed CATT, has the following configurations. The audio encoder is a 12-layer transformer (in the 14-million-parameter model case, denoted as 14M below) or a 20-layer transformer (in the 22-million-parameter model case, denoted as 22M below). The label encoder is a 4-layer transformer (in 14M case) or an 18-layer transformer (in 22M case). Both the audio and label encoders have an embedding size of 256 ($d_a$ and $d_l$ in Section~\ref{sec:tt}) with 4 attention heads. The joint network is a fully-connected feed-forward component with one hidden layer followed by a $\tanh$ activation function. 

We explore two context embedding extractors: (1) 1-layer and 2-layer BLSTM of 256 dimensions ($d_c$ in Section~\ref{sec:catt}) for the 14M model and 22M model, respectively; (2) 4-layer, 14M SmallBERT \cite{turc2019} of 256 dimensions. The context biasing encoder consists of a 4 transformer blocks with an embedding size of 256  ($d=d_{ca}$ in Section~\ref{sec:catt}) and 4 attention heads; all of the above transformer blocks also contain layer normalizations, feed-forward layers after attention layers and residual links in between layers; there are 512 neurons for the feed-forward layers. The baseline models, i.e. transformer transducers, do not have context embedding extractors or context biasing encoders. We train baselines and CATT models from scratch, except for the SmallBERT context encoder.

The input audio features fed into the network consists of 64-dimensional LFBE features, which are extracted every 10 ms with a window size of 25 ms from audio samples. The features of each frame is then stacked with the left two frames, followed by a downsampling of factor 3 to achieve low frame rate, resulting in 192 feature dimensions. 
The subword tokenizer \cite{sennrich-etal-2016-neural, kudo2018subword} is used to create tokens from the transcriptions; we use 4000 subwords/tokens in total. 
We trained the models by minimizing the alignment loss~\cite{graves2012sequence} using
the Adam optimizer \cite{kingma2014adam}, and varied the learning rate following \cite{Dong2018SpeechTransformerAN,NIPS2017_3f5ee243}. 

%% file: results_and_discussions.tex
\section{Results and Discussion}\label{sec:results}

\subsection{Comparisons to the Non-contextual Baselines and Shallow Fusion}
The results of all experiments are demonstrated as the relative word error rate reduction (WERR). Given a model A's WER ($\text{WER}_A$) and a baseline B's WER ($\text{WER}_B$), the WERR of A over B is computed as
$
\text{WERR} = (\text{WER}_B - \text{WER}_A)/\text{WER}_B.
$
In the following experiments, we use a vanilla transformer transducer (T-T) as our baseline. We denote shallow fusion and context embedding as SF and CE, respectively. We compare the proposed CATT with the following models: (1) T-T, (2) T-T + SF, (3) CATT (BLSTM-CE), (4) CATT (BLSTM-CE) + SF, (5) CATT (SmallBERT-CE), (6) CATT (SmallBERT-CE) + SF. We freeze the SmallBERT model during training, and shallow fusion is only applied during inference time.

Table~\ref{tab:14M_auto_WER} presents the WERRs of the proposed CATT over the baselines. We can see that CATT outperforms both T-T and SF on both \emph{Personalized} and \emph{Common} sets. The reason CATT also improves the WERs on \emph{Common} sets is that contextual phrases (e.g. the device location and device setting) may appear in the utterances of the \emph{Common} set.
Specifically, we can see that CATT with BLSTM context embeddings outperforms the vanilla transformer transducer by up to 2.2\%. If we leverage a pre-trained SmallBERT as the context encoder, the CATT model outperforms shallow fusion with 4.7\% vs 1.6\% WERR on \emph{Personalized} test sets. Combining CATT and SF further achieves 6.3\% and 6.5\% WERR on both \emph{Personalized} and \emph{Common} test sets.

Note that the SmallBERT encoder has around 14M non-trainable parameters, which is about the same size as the trainable part of the baseline model in Table~\ref{tab:14M_auto_WER}. To fully take advantage of the strong semantic prior knowledge of BERT, we use a larger baseline T-T in Table~\ref{tab:35M_auto} with 22M trainable parameters. We get 10.6\% and 10.3\% WERR on \emph{Personalized} and \emph{Common} test sets for CATT with BLSTM context embeddings. CATT with the SmallBERT context embeddings further improves the WERR to 15.3\% on \emph{Personalized} test sets. Besides, CATT performs better than shallow fusion with 15.3\% vs 3\% WERR on \emph{Personalized} test sets. One drawback of shallow fusion is that it is very sensitive to the boosting weights leading to over boosting and degrades the WERs. Instead, CATT consistently improves WERs over the baselines.

\begin{figure*}[t]
  \centering
  \includegraphics[width=0.8\linewidth]{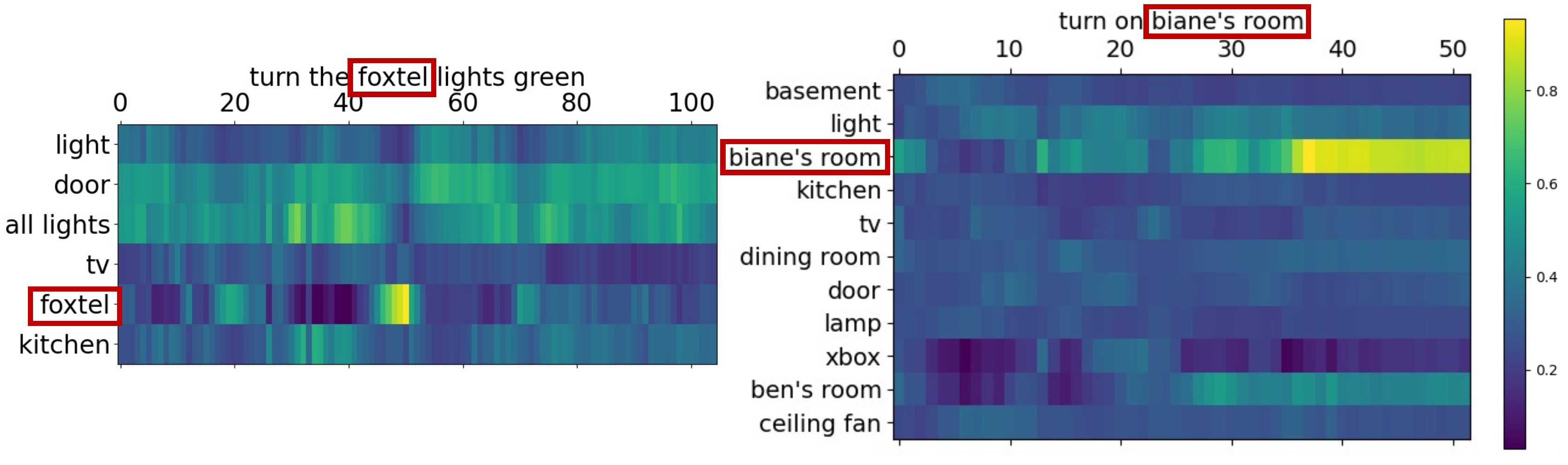}
  \vskip -8pt
  \caption{Visualizing attention weights from Eqn.(\ref{eq:cc}) for two examples with entity names which are marked by red-colored rectangles. The $x$-axis shows the audio frame indices and the corresponding predictions, and the $y$-axis shows the context phrases associated with the example.  While the darker colors represent values close to zero, the brighter ones represent higher weights.}
  \label{fig:attention}
\end{figure*}

Two example outputs generated by T-T and CATT are shown in Table~\ref{predictions}. We also visualize the attention values in Fig.~\ref{fig:attention}, to better understand how the CATT attention mechanism works. As can be seen, the CATT attends to the entities and give them higher attention weights. 

\begin{table}[t]
\footnotesize
\caption{Comparing outputs generated by the T-T baseline and CATT. Entity names and personalized device names are highlighted with bold font in these examples.}
\vskip 3pt
\centering
\label{predictions}
\begin{tabular}{|l|l|l|l|l|}
\hline
\textbf{Method}        & \textbf{Output} \\ \hline
Reference Utterance       & turn on \textbf{baine's} room                      \\ \hline
T-T & turn on basement room  \\\hline
CATT & turn on \textbf{baine's} room \\\hline\hline
Reference Utterance        & turn the \textbf{foxtel} lights green                     \\ \hline
T-T  & turn the fox lights green    \\\hline
CATT & turn the \textbf{foxtel} lights green     \\\hline
\end{tabular}
\end{table}

\begin{table}[t]
\footnotesize
\caption{WERRs for 22M CATT vs.T-T and SF}
\vskip 3pt
\centering
\label{tab:35M_auto}
\begin{tabular}{|l|c|c|c|c|}
\hline
 & \multicolumn{2}{c|}{\emph{Personalized}}                        & \multicolumn{2}{c|}{\emph{Common}}                          \\ \hline
WERR (\%)    & \multicolumn{1}{l|}{dev} & \multicolumn{1}{l|}{test} & \multicolumn{1}{l|}{dev} & \multicolumn{1}{l|}{test} \\ \hline
T-T                     & 0 & 0 & 0 & 0  \\ \hline
T-T + SF                & 4.1     & 3.0     & 2.0     &  2.1    \\ \hline \hline
CATT (BLSTM-CE)   & 11.0 & 10.6 & 10.8 & 10.3 \\ \hline
CATT (BLSTM-CE) + SF     & 12.3     &10.8      & 12.8     &  12.4    \\ \hline \hline
CATT (SmallBERT-CE) & 15.1     &13.6     & 13.7     &  12.4    \\ \hline
CATT (SmallBERT-CE) + SF    &16.6   &  15.3    &  14.7    &   13.4        \\ \hline
\end{tabular}
\end{table}

\begin{figure}[t]
  \centering
  \includegraphics[width=0.7\linewidth]{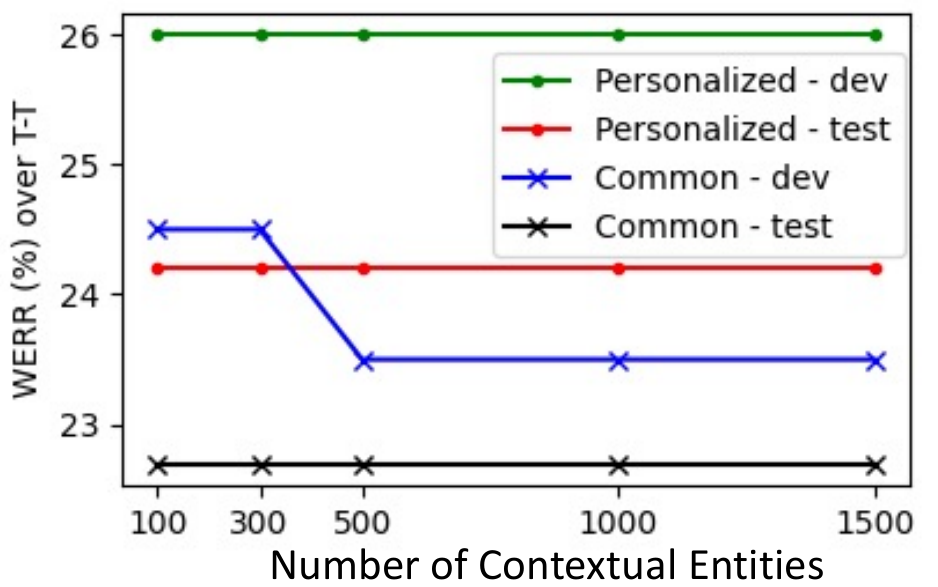}
  \vskip -8pt
  \caption{WERRs (\%) for CATT over T-T with different number of contextual entities $K$}
  \label{fig:catalog_size}
\end{figure}

\subsection{Comparisons of Different Queries to Attend Context}
Given the best performed results of CATT with SmallBERT context embeddings in the last section, we take this model and further evaluate it by using the audio embedding as queries (Fig.~\ref{fig:catt_diagram} (c)) and both audio and label embeddings as queries (Fig.~\ref{fig:catt_diagram} (d)) to attend the context.  They are denoted as audio-Q and audio+label-Q respectively as follows. In Table~\ref{tab:14M_auto_WER_audioLabel_query} and Table~\ref{tab:35M_auto_audioLabel_query}, we see consistent improvements across different model sizes when using both audio embeddings and label embeddings to attend context. We believe that context embeddings have a correlation with both audio and label embeddings. In addition, we also hypothesize that better alignment can be learned between input audio and output token predictions when both embeddings are calibrated via context.
\begin{table}[t]
\footnotesize
\caption{WERRs for 14M CATT with audio query and audio+label query}
\vskip 3pt
\centering
\label{tab:14M_auto_WER_audioLabel_query}
\begin{tabular}{|l|c|c|c|c|}
\hline
  & \multicolumn{2}{c|}{\emph{Personalized}}                        & \multicolumn{2}{c|}{\emph{Common}}                          \\ \hline
WERR (\%)    & \multicolumn{1}{l|}{dev} & \multicolumn{1}{l|}{test} & \multicolumn{1}{l|}{dev} & \multicolumn{1}{l|}{test} \\ \hline
T-T                     & 0 & 0 & 0 & 0  \\ \hline
CATT(SmallBERT-CE),audio-Q    &  4.3    &  4.7     & 4.1     & 4.3     \\ \hline
CATT(SmallBERT-CE),audio+label-Q  & 11.4  & 10.9 &   9.3   &  9.7  \\ \hline
\end{tabular}
\end{table}
\vskip -10pt

\begin{table}[t]
\footnotesize
\caption{WERRs for 22M CATT with audio query and audio+label query}
\vskip 3pt
\centering
\label{tab:35M_auto_audioLabel_query}
\begin{tabular}{|l|c|c|c|c|}
\hline
 & \multicolumn{2}{c|}{\emph{Personalized}}                        & \multicolumn{2}{c|}{\emph{Common}}                          \\ \hline
WERR (\%)    & \multicolumn{1}{l|}{dev} & \multicolumn{1}{l|}{test} & \multicolumn{1}{l|}{dev} & \multicolumn{1}{l|}{test} \\ \hline
T-T                     & 0 & 0 & 0 & 0  \\ \hline
CATT(SmallBERT-CE),audio-Q & 15.1     &13.6     & 13.7     &  12.4    \\ \hline
CATT(SmallBERT-CE),audio+label-Q     & 26.0  & 24.2 &  24.5 & 22.7        \\ \hline
\end{tabular}
\end{table}

\subsection{CATT with Varying Number of Contextual Entities}
As mentioned in Section~\ref{sec:dataset2}, we randomly add less relevant contexts in addition to the relevant ones for the model robustness to the irrelevant contexts. To verify it, we evaluate the best-performing CATT in Table~\ref{tab:35M_auto_audioLabel_query} with different number of contextual phrases, $K$, chosen randomly from the complete context list, and we assume the target context is present in the list. As can be seen in Fig.~\ref{fig:catalog_size}, the WERRs of CATT over T-T remain across different test sets with different values of $K$. 

\subsection{Comparisons to Contextual LAS}
Finally, we compare the proposed CATT with an existing deep contextual model -- Contextual LAS (C-LAS)~\cite{pundak2018deep}, which was also designed to bias towards the personalized words during training and inference. We follow \cite{pundak2018deep} to set up the configurations of audio encoder, context encoder, and the attention module, except that the number of hidden units are adjusted to match our model sizes (14M and 22M parameters). The details of C-LAS configuration we used are described below (The number before ``/'' is for 14M model while the number after ``/'' is for 22M case): The audio encoder’s architecture consists of 10 BLSTM layers, each with 128 / 224 nodes. The audio encoder attention is computed over 256 / 64 dimensions, using 4 attention heads. The bias-encoder consists of a single BLSTM layer with 256 nodes, and the bias-attention is computed over 256 / 64 dimensions. Finally, the decoder consists of 4 LSTM layers with 256 / 128 nodes. The WERR results of CATT over C-LAS are presented in Table~\ref{tab:14M_auto_WER_vs_CLAS} and Table~\ref{tab:22M_auto_WER_vs_CLAS} for 14M and 22M parameters individually.

With the same type of context encoder, a BLSTM, the proposed CATT has shown $\sim 3$-$6\%$ relative WER improvements. The CATT can be further improved by exploiting the BERT based context embedding and achieve up to $8.1\%$ relative improvement in 22M case with an audio query. Again, CATT with both audio and label embeddings as queries (audio+label-Q) performs the best (by up to $21.7\%$ relative improvement against C-LAS for 22M model).

\begin{table}[th]
\footnotesize
\caption{WERRs for 14M CATT vs.C-LAS}
\vskip 3pt
\centering
\label{tab:14M_auto_WER_vs_CLAS}
\begin{tabular}{|l|c|c|c|c|}
\hline
  & \multicolumn{2}{c|}{\emph{Personalized}}                        & \multicolumn{2}{c|}{\emph{Common}}                          \\ \hline
WERR (\%)    & \multicolumn{1}{l|}{dev} & \multicolumn{1}{l|}{test} & \multicolumn{1}{l|}{dev} & \multicolumn{1}{l|}{test} \\ \hline
C-LAS~\cite{pundak2018deep} & 0 & 0 & 0 & 0  \\ \hline
CATT(BLSTM-CE),audio-Q  & 4.2 & 4.6 & 4.0 & 3.2  \\ \hline
CATT(SmallBERT-CE),audio-Q   &  6.9  & 7.6 &  6.1  & 5.3     \\ \hline
CATT(SmallBERT-CE),audio+label-Q   & 13.9 & 13.6 &  11.1  & 10.6    \\ \hline
\end{tabular}
\end{table}

\begin{table}[ht!]
\footnotesize
\caption{WERRs for 22M CATT vs.C-LAS}
\vskip 3pt
\centering
\label{tab:22M_auto_WER_vs_CLAS}
\begin{tabular}{|l|c|c|c|c|}
\hline
  & \multicolumn{2}{c|}{\emph{Personalized}}                        & \multicolumn{2}{c|}{\emph{Common}}                          \\ \hline
WERR (\%)    & \multicolumn{1}{l|}{dev} & \multicolumn{1}{l|}{test} & \multicolumn{1}{l|}{dev} & \multicolumn{1}{l|}{test} \\ \hline
C-LAS~\cite{pundak2018deep} & 0 & 0 & 0 & 0  \\ \hline
CATT(BLSTM-CE),audio-Q    &  5.8 &  4.8 &  2.2 &  1.1 \\ \hline
CATT(SmallBERT-CE),audio-Q   &   10.1   & 8.1  & 5.4  & 3.4    \\ \hline
CATT(SmallBERT-CE),audio+label-Q   &  21.7    &  19.4   &  17.2  & 14.8    \\ \hline
\end{tabular}
\end{table}

%% file: conclusion.tex
\section{Conclusion}
We proposed a novel CATT model to enable a state-of-the-art transformer transducer ASR model to take advantage of the contextual data both in training and inference. We leveraged both BLSTM and BERT-based models to encode context. The relevance of context was measured by the proposed multi-head cross-attention mechanism with audio embeddings alone or together with label embeddings. Our experiments showed that CATT outperformed the non-contextual models, shallow fusion, and C-LAS models on an in-house dataset with a variety of named entities and personalized information.

%% file: acknowledgement.tex
\noindent\textbf{Acknowledgement}
\ninept
We thank Christian Hensel, Anirudh Raju, Zhe Zhang, G. Tiwari, Grant P. Strimel, T. Muniyappa, Kanthashree Sathyendra, Kai Wei, and Markus Mueller for valuable feedbacks.